\documentclass[conference]{IEEEtran}
\IEEEoverridecommandlockouts
\usepackage{cite}
\usepackage{amsmath,amssymb,amsfonts}
\usepackage{algorithmic}
\usepackage{graphicx}
\usepackage{textcomp}
\usepackage{xcolor}

\usepackage{times}

\usepackage{soul}
\usepackage{url}
\usepackage[hidelinks]{hyperref}
\usepackage[utf8]{inputenc}
\usepackage[small]{caption}
\usepackage{booktabs}
\usepackage{multirow}
\usepackage{caption}
\usepackage{float}
\usepackage{todonotes}
\usepackage{balance}
\usepackage{fancyhdr}

\usepackage{acronym}
\usepackage{amsmath}
\usepackage{amssymb}

\acrodef{LIME}[LIME]{Local Interpretable Model-agnostic Explanations}
\acrodef{DL}[DL]{Deep-Learning}
\acrodef{ML}[ML]{Machine Learning}
\acrodef{XAI}[XAI]{eXplainable Artificial Intelligence}
\acrodef{SHAP}[SHAP]{SHapley Additive exPlanations}
\acrodef{Grad-CAM}[Grad-CAM]{Gradient-weighted Class Activation Mapping}
\acrodef{CNN}[CNN]{Convolutional Neural Network}
\acrodef{AI}[AI]{Artificial Intelligence}
\acrodef{XUI}[XUI]{eXplanation User Interface}
\acrodef{FRBS}[FRBS]{Fuzzy Rule-Based Systems}

\def\BibTeX{{\rm B\kern-.05em{\sc i\kern-.025em b}\kern-.08em
    T\kern-.1667em\lower.7ex\hbox{E}\kern-.125emX}}

\begin{document}

\title{A Theoretical Framework for AI Models Explainability with Application in Biomedicine}

\author{\IEEEauthorblockN{Matteo Rizzo}
\IEEEauthorblockA{
\textit{Department of Computer Science} \\
\textit{Ca' Foscari University}\\
Venice, Italy \\
matteo.rizzo@unive.it}
\and
\IEEEauthorblockN{Alberto Veneri}
\IEEEauthorblockA{\textit{Department of Computer Science} \\
\textit{Ca' Foscari University \& ISTI-CNR}\\
Venice, Italy \\
alberto.veneri@unive.it}
\and
\IEEEauthorblockN{Andrea Albarelli}
\IEEEauthorblockA{\textit{Department of Computer Science} \\
\textit{Ca' Foscari University}\\
Venice, Italy \\
albarelli@unive.it}
\and
\IEEEauthorblockN{Claudio Lucchese}
\IEEEauthorblockA{\textit{Department of Computer Science} \\
\textit{Ca' Foscari University}\\
Venice, Italy \\
claudio.lucchese@unive.it}
\and
\IEEEauthorblockN{Marco Nobile}
\IEEEauthorblockA{\textit{Department of Computer Science} \\
\textit{Ca' Foscari University}\\
Venice, Italy \\
marco.nobile@unive.it}
\and
\IEEEauthorblockN{Cristina Conati}
\IEEEauthorblockA{\textit{Department of Computer Science} \\
\textit{University of British Columbia}\\
Vancouver, Canada \\
conati@ubc.cs.ca}
}

\maketitle


\begin{abstract}
\Ac{XAI} is a vibrant research topic in the artificial intelligence community.
It is raising growing interest across methods and domains, especially those involving high stake decision-making, such as the biomedical sector. 
Much has been written about the subject, yet \ac{XAI} still lacks shared terminology and a framework capable of providing structural soundness to explanations. 
In our work, we address these issues by proposing a novel definition of explanation that synthesizes what can be found in the literature. We recognize that explanations are not atomic but the combination of evidence stemming from the model and its input-output mapping, and the human interpretation of this evidence. Furthermore, we fit explanations into the properties of faithfulness (\textit{i.e.}, the explanation is an accurate description of the model's inner workings and decision-making process) and plausibility (\textit{i.e.}, how much the explanation seems convincing to the user). 
Our theoretical framework simplifies how these properties are operationalized, and it provides new insights into common explanation methods that we analyze as case studies.  
We also discuss the impact that our framework could have in biomedicine, a very sensitive application domain where XAI can have a central role in generating trust.
\end{abstract}


\section{Introduction}
\label{sec:introduction}

The advent of \ac{DL} allowed for raising the accuracy bar of \ac{ML} models for countless tasks and domains. Riding the wave of enthusiasm around such stunning results, \ac{DL} models have been deployed even in high-stake decision-making environments, not without criticism~\cite{Rudin2019-bj}. 
These kinds of environments require not only high predictive accuracy but also an \emph{explanation} of why that prediction was made. The need for explanations initiated the discussion around the explainability of \ac{DL} models, which are known to be ``black boxes". In other words, their inner workings are hard for humans to be understood. Who should be accountable for a model-based decision and how a model came to a certain prediction are just some of the questions that drive research on explaining \ac{ML} models. 
This is particularly relevant, for instance, in the biomedical field, where human lives are at stake, and understanding the reasoning behind a model's predictions is essential to guarantee safety and avoid costly errors~\cite{Kundu2021}. Furthermore, the ability to explain how a model arrived at a certain conclusion can increase the understanding of the underlying biological mechanisms, enabling more informed support to decision-making for clinicians and researchers.
With the first recent attempts of the legislative machinery to make explanations for automatic decisions a user’s right~\cite{Goodman2017-iw}, the pressure on generating explanations for the \ac{ML} model's behaviors raised even more. 
Despite the endeavor of the \ac{XAI} community to develop either models that are explainable by design~\cite{Chen2018-eq,Zhang2017-by,Hou2020-zf} and methods to explain existing black-box models~\cite{Ribeiro2016-uy,Lundberg2017-ar,Bahdanau2014-fv}, the way to \ac{DL} explainability is paved with results that are mostly preliminary and anecdotal in nature (\textit{e.g.},~\cite{Jain2019-oe,Wiegreffe2019-ht,Serrano2019-tm}). 
Most notably, it is hard to relate different pieces of research due to a lack of common theoretical grounds capable of supporting and guiding the discussion. 
In particular, we detect a gap in the literature on foundational issues such as a shared definition of the term ``explanation" and the users' role in the design and deployment of explainability for complex \ac{ML} models. The XAI community suffers from the paucity of common terminology, with only a few attempts of establishing one, focusing more on the distinction among the terms ``interpretable", ``explainable", and ``transparent" rather than the inner structure and meaning of an explanation (\textit{e.g.},~\cite{Graziani,clinciu-hastie-2019-survey,murdoch_definitions_2019}). 
Similarly, a lack of an outline of the main theoretical components of the discussion around explainability disperses research, while the current literature finds it hard to provide the involved stakeholders with principled analytical tools to operate on black-box models. This trend has been detected in the delicate field of biomedicine and addressed with context-specific guidelines~\cite{Arbelaez_Ossa2022-ve}.  
In this work, we propose a simple, general, and effective theoretical framework that outlines the core components of the explainability machinery and lays out grounds for a more coherent debate on how to explain the decisions of \ac{ML} models. Such a framework is not meant to be set in stone but rather to be used as a common reference among researchers and iteratively improved to fit more and more sophisticated explainability methods and strategies. We hope to provide shared jargon and formal definitions to inform and standardize the discussion around crucial topics of \ac{XAI}. 
The core of the proposed theoretical framework is a novel definition of explanation, that draws from existing literature in sociology and philosophy but, at the same time, is easy to operationalize when analyzing a specific approach to explain the predictions made by a model. 
We conceive an explanation as the interaction of two decoupled components, namely \emph{evidence} and its \emph{interpretation}. 
Evidence is any sort of information stemming from a \ac{ML} model, while an interpretation is some semantic meaning that human stakeholders attribute to the evidence to make sense of the model’s inner workings.
We relate these definitions to crucial properties of explanations, especially \emph{faithfulness} and \emph{plausibility}.
Jacovi \& Goldberg define faithfulness as ``the accurate representation of the causal chain of decision-making processes in a model"~\cite{Jacovi2020-ec}. 
We argue that faithfulness relates in different ways to the elements of the proposed theoretical framework because it assures the interpretation of the evidence is true to how the model actually uses it within its inner reasoning. 
A property orthogonal to faithfulness is plausibility, namely ``the degree to which some explanation is aligned with the user’s understanding of the model's decision process" \cite{Jacovi2020-ec}. 
A follow-up work by Jacovi \& Goldberg addresses plausibility as the ``property of an explanation of being convincing towards the model prediction, regardless of whether the model was correct or whether the interpretation is faithful"~\cite{Jacovi2021-pi}. 
We relate plausibility to faithfulness and highlight a need for faithfulness to be embedded in explainability methods and strategies, as well as plausibility as an important (yet not indispensable) property of the same. This is particularly true in the biomedical field as a high stake environment, where it is crucial for the explanation to portray the decision-making process of the model accurately.
As case studies, we zoom in on the evaluation of faithfulness of some popular \ac{DL} explanation tools and strategies, such as ``attention"~\cite{Bahdanau2014-fv,Vaswani2017-kq}, \ac{Grad-CAM}~\cite{Selvaraju2016-hw} and \ac{SHAP}~\cite{Lundberg2017-ar}. In addition, we look at the faithfulness of models traditionally considered intrinsically interpretable (a notion with distance ourselves to) such as a \emph{linear regressors} and models based on \emph{fuzzy logic}.

\section{Designing Explainability}
\label{sec:designing-explainability}

Research in \ac{XAI} seizes the problem of explaining models for decision-making from multiple perspectives. 
First of all, we observe that most of the existing literature uses the terms ``interpretable" and ``explainable" interchangeably, while some have highlighted the semantic nuance that distinguishes the two words~\cite{Mittelstadt2019-jk}. 
We argue that the term \emph{explainable} (and, by extension \emph{explainability}) is more suited than the term \emph{interpretable} (similarly, by extension, \emph{interpretability}) to describe the property of a model for which effort is made to generate human-understandable clarifications of its decision-making process. 
The definition of \emph{explanation} is thus crucial and will be discussed extensively in \autoref{sec:defining-explanations}.
Our claim follows two rationales: \emph{(i)} the term \emph{interpretation} is used within our proposed framework with a precise meaning that deviates from the current literature and that we deem more accurate (see \autoref{sec:interpretation}); \emph{(ii)} we argue against grouping models into \emph{inherently interpretable} and \emph{post-hoc explainable}. 
Recently, Molnar has defined ``intrinsic interpretability" as a property of \ac{ML} models that are considered fully understandable due to their simple structure (\textit{e.g.}, short decision trees or sparse linear models), while ``post hoc explainability" as the need for some models to apply interpretation methods after training~\cite{Molnar2022-kh}.
Although principled, we drop this hard distinction by claiming that all models embed a certain degree of explainability.
Even though, to the best of our knowledge, no metrics can quantify explainability yet, we can assert that this depends on multiple factors. 
In particular, a model is as explainable as the explanations that are proposed to the user to justify a certain prediction are effective. 
Thus, bringing the human into the explainability design loop is key to deploying models that are actually explainable. 
Consequently, there are models for which it is easier to design explanations (\textit{i.e.}, the so-called \emph{white-box} models, \textit{e.g.}, linear regression, decision trees, rule-based systems, etc.) and models for which the same process is more difficult (\textit{i.e.}, the so-called \emph{black-box} models, \textit{e.g.}, artificial neural networks). 
The notion of difficulty here is defined by the inner complexity of the model, which relates to the amount of cognitive load the user can sustain. 
We highlight that the degree of explainability moves along a gradient from black-box to white-box models, without clear-cut thresholds. 
Nevertheless, in \autoref{sec:case-studies}, we show that explanations for both white-box and black-box models fit our proposed framework. 
Thus they both can be structured homogeneously and more deeply understood by leveraging theoretical tools. 
Most importantly, we advocate for explainability design as a crucial part of \ac{AI} software development. 
We endorse Chazette et al., in claiming that explainability should be considered a non-functional requirement in the software design process~\cite{Chazette2020-zz}. 
Thus explanations for any \ac{ML} models (and, especially, for \ac{DL} models) should be accounted for within the initial design of an \ac{AI}-powered application. Even the most accurate black-box model should not be deployed without an explanation mechanism backing it up, as we cannot be sure whether it learned to discriminate over meaningful or wrong features.  A classic example is a dog image classification model learning to detect huskies because of the snowy setting instead of the features of the animal itself, involuntarily deceiving the users~\cite{Ribeiro2016-uy}. A design-oriented approach to \ac{AI} development should involve taking humans into the loop, thus fostering a human-centered \ac{AI} which is more intelligible by design and is expected to increase trust in the end-users~\cite{Li2022-xt}.

\section{Characterising the Inference Process of a Machine Learning Model}

In this section, we provide a formal characterization of the inference process of a general \ac{ML} model, without any constraint on the task. Such a characterization will be used to introduce the terminology which substantiates the main components of our proposed framework of explainability, whose details are provided in \autoref{sec:defining-explanations}. To this end, we define a \ac{ML} model $M$ as an arbitrarily complex function mapping a \emph{model input} to a \emph{model conclusion} through a sequential composition of \emph{transformation steps}. The whole characterization is exemplified in~\autoref{fig:transformation-functions}.

\subsection{Elements of the Characterization}

\noindent
\textbf{Model Input.} The model input consists of a set of features, either coming from an observation or synthetically generated. 

\noindent
\textbf{Model Conclusion.} The model conclusion is the final output of the model, which is the outcome of the last link in the chain of transformations over the model input.

\noindent
\textbf{Transformation steps.} Overall, the decision-making process of $M$ can be represented as a chain of $N > 0$ transformations of the original model input, that are causally related. This causal chain is enforced by model design (\textit{e.g.}, the sequence of layers in the architecture of a neural network or the depth of a decision tree). We call each stage of this causal chain a ``transformation step", and we denote it with $s_i$, for $i\in [1, N]$. The transformation steps advance the computation from the model input to the model output through \emph{transformation functions}.

\begin{figure}[t]
    \centering
    \includegraphics[width=0.45\textwidth]{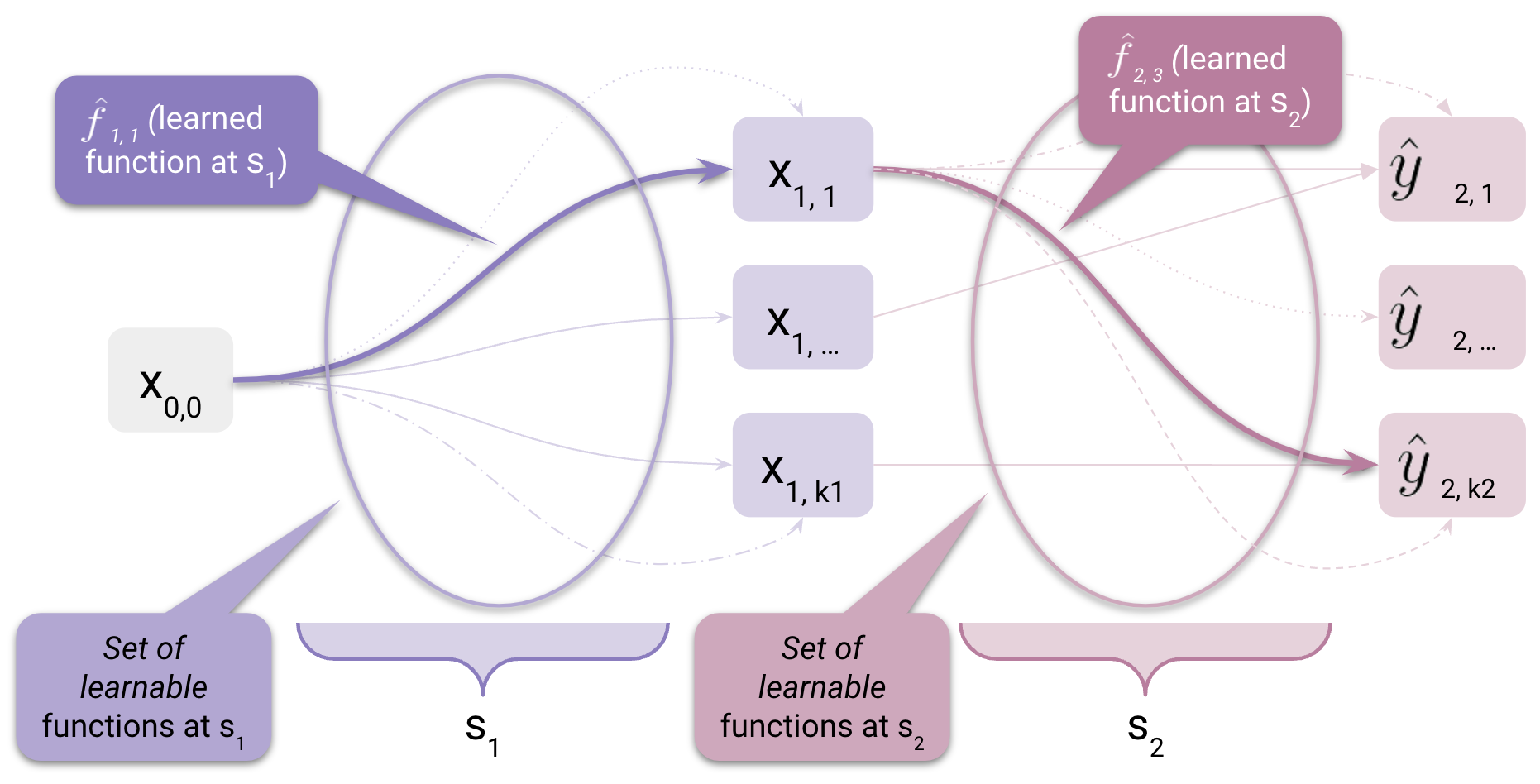}
    \caption{Example of transformation functions for two steps $s_i$.}
    \label{fig:transformation-functions}
\end{figure}

\noindent
\textbf{Transformation functions.} Each transformation step $s_i$ relates to a set of $n_i$ ``transformation functions" $f_{i,m_i}$, where $m_i\in[1,n_i]$ indicates one of the possible learnable functions at $s_i$. Note that, in general, the number of such functions would be infinite, but we discretize it assuming that we are working on a real scenario using some computational machine. The transformation functions are mappings from a feature set $x_{i-1,j}$ to a feature set $x_{i,z}$, with $j\in [1, k_{i-1}]$, $z\in [1, k_i]$ (\textit{i.e.}, the arrows enclosed in the ellipses in \autoref{fig:transformation-functions}). The number $k_i$ denotes the cardinality of the set of all possible feature sets generated by all possible learnable transformation functions at step $s_i$. These transformation functions are generally opaque to the user in the context of the so-called black-box models. At every step in the chain of transformation steps, the model learns one of the possible transformation functions (\textit{i.e.}, the optimal function according to some learning scheme, highlighted with a solid line in \autoref{fig:transformation-functions}). That is, the model learns the function $\hat{f}_{i,m_i}$ such that $\hat{f}=\hat{f}_{N,m_N} \circ \ldots \circ\hat{f}_{i,m_i} \circ \ldots \circ \hat{f}_{1,m_1}$ is the overall approximation of the true mapping from the model input to the model conclusion. According to the notation above, we denote the model input as $x_{0,0}$ (or simply $x$) and the model conclusion as $\hat{y}_{N,j}$, with $j\in [1, k_N]$ (or simply $\hat{y}$). 

\subsection{Observations}
\noindent
We asserted that, at each transformation step $s_i$, the model picks one function $\hat{f}_{i, m_i}$ among $n_i$ such that $\hat{f}_{i, m_i}(x_{i-1},j)=x_{i,z}$. This raises issues that increase model opacity. At step $s_i$ the chosen function $\hat{f}_{i, m_i}$ can map different intermediate transformations $x_{i-1,j}$ of the feature set at the previous transformation step into the same transformation $x_{i,z}$ one step further in the chain. This means that the same outcome in the transformation chain, be it intermediate or conclusive, can be achieved through different rationales, and it could be difficult for a human user to understand which of them is the one the model has actually learned. This can be a result of a high dimensionality of the set of transformation functions, as well as a high complexity of the transformed feature set. 

For example, pictures of zebras and salmon can be discriminated on the basis of either their anatomy (i.e., zebras have stripes, salmon have gills) or the environment/habitat (i.e., zebras live in savannas and salmon in rivers). If we consider a relatively complex model such as a \ac{CNN}, where a transformation step coincides with a layer within the network architecture, it is generally difficult to understand which kind of transformation $f_{i,m_i}$ this represents, if any that is human-understandable. Thus, how do we make sense of which of the $n_i$ possible alternative mappings of $x_{i-1,j}$ led to $x_{i,z}$? This remains an open question, with major implications for the discussion around faithfulness, which we will enlarge in the next section.

\section{Defining explanations}
\label{sec:defining-explanations}

Recent work on \ac{ML} interpretability produced multiple definitions for the term ``explanation". According to Lipton, ``explanation refers to numerous ways of exchanging information about a phenomenon, in this case, the functionality of a model or the rationale and criteria for a decision, to different stakeholders"~\cite{Lipton2016-ba}. Similarly, for Guidotti et al. ``an explanation is an ``interface" between humans and a decision-maker that is at the same time both an accurate proxy of the decision-maker and comprehensible to humans"~\cite{Guidotti2018-ti}. Murdoch et al. add to how the explanation is delivered to the user stating that ``an explanation is some relevant knowledge extracted from a machine-learning model concerning relationships either contained in data or learned by the model. [...] They can be produced in formats such as visualizations, natural language, or mathematical equations, depending on the context and audience"~\cite {Murdoch2019-wk}. On a more general note, Mueller et al. state that
``the property of ``being an explanation" is not a property of the text, statements, narratives, diagrams, or other forms of material. It is an interaction of (i) the offered explanation, (ii) the learner’s knowledge and beliefs, (iii) the context or situation and its immediate demands, and (iv) the learner’s goals or purposes in that context"~\cite{Mueller2019-jr}. Finally, Miller tackles the challenge of defining explanations from a sociological perspective. The author highlights a wide taxonomy of explanations but focuses on those which are an answer to a ``why-question" \cite{Miller2017-wj}.

\begin{figure}[b]
    \centering
    \includegraphics[width=0.49\textwidth]{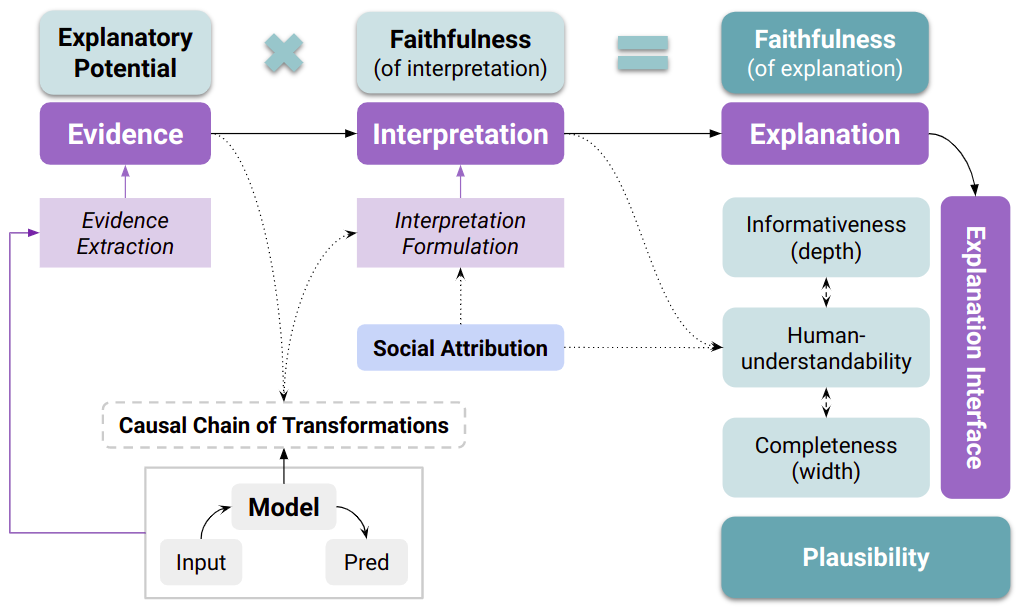}
    \caption{Overview of the theoretical framework of explainability.}
    \label{fig:framework}
\end{figure}

The definitions mentioned above offer a well-rounded perspective on what constitutes an explanation. However, they fail to highlight its atomic components and to characterize their relationships. We synthesize our proposed definition of explanation based on complementary aspects of the existing definitions. The result is a concise definition that is easy to operationalize for supporting the analysis of multiple approaches to explainability. Our full proposed framework is reported in the scheme in \autoref{fig:framework}, whose components will be discussed in the following sections. 

\subsection{Explanation}
\label{sec:explanation}

Given a model $M$ which takes an input $x$ and returns a prediction $\hat{y}$, we define \emph{explanation} the output of an \emph{interpretation function} applied to some \emph{evidence}, providing the answer to a ``why question'' posed by the user.

\subsection{Evidence}
\label{sec:evidence}

\emph{Evidence} ($e$) is whatever kind of objective information stemming from the model we wish to provide an explanation for and that can reveal insights into its inner workings and rationale for prediction (\textit{e.g.}, attention weights, model parameters, gradients, etc.).

\subsubsection{Evidence Extractor}
\label{sec:evidence-extractor}

An \emph{evidence extractor} ($\xi$) is a method fetching some relevant information about either $M$, $x$, $\hat{y}$, or a combination of the three. Then: $ e = \xi(x, \hat{y}, M)$. Examples of evidence extractors are, e.g., encoder plus attention layers, gradient back-propagation, and random tree approximation, with the corresponding extracted evidence being attention weights, gradient values, and random tree mimicking the original model. In the peculiar case of a white-box approach, that is, ML models designed to be \emph{easily interpretable} by the user (e.g., linear regression, fuzzy rule-based systems) the extraction of evidence is straightforward since all components of the model directly present a piece of semantic information in a human-comprehensible format. 

\subsubsection{Explanatory Potential}
\label{sec:explanatory-potential}

We define \emph{explanatory potential} ($\epsilon(e)$) of some evidence as the extent to which the evidence influences the causal chain of transformations steps of a model. Intuitively, the explanatory potential indicates ``how much'' of a model the selected type of evidence can explain. It can be computed either by counting how many transformation steps are impacted by the evidence (i.e., \textit{breadth}), or how much of each single transformation step is impacted by the evidence (i.e., \textit{depth}).

\subsection{Interpretation}
\label{sec:interpretation}

An \emph{interpretation} is a function $g$ associating semantic meaning to some evidence and mapping its instances into explanations either for a given prediction or the whole model. 
Then an explanation can be defined as either $E=g(e, x, \hat{y}, M)$, or $E=g(e, M)$, respectively.

\subsubsection{Local vs. Global Interpretations}
\label{sec:local-vs-global}

In accordance with the existing literature, we relate ``evidence" and ``interpretation" to the concepts of \textit{locality} and \textit{globality}. Both evidence and interpretations can either be local or global. Local evidence (e.g., attention weights, gradient, etc.) relates relevant model information to a particular model input $x$ and corresponding prediction $\hat{y}$. Global evidence (e.g. full set of model parameters) is generally independent of specific inputs and might explain higher level functioning (providing deeper or wider info) of the model or some of its sub-components. Similarly, interpretations can provide either a local or a global semantic of the evidence. A local interpretation of attention could be, \textit{e.g.}, ``attention weights are descriptive of input components’ importance to model output". On the other hand, a global interpretation of the same evidence may aggregate all the attention weights' heatmaps for a whole dataset and highlight specific patterns. For example, in a dog vs. cat classification problem, a global interpretation of attention may be represented by clusters of similar parts of the animal's body (\textit{e.g.}, groups of ears, tails, etc.) highlighted by the attention activations.

\subsubsection{Generating Interpretations}
\label{sec:generating-interpretations}

Given some evidence involved in one or more steps $s_i$ of $M$, we guess how this evidence is involved in the opaque input-to-output transformations by formulating an interpretation $g$ of some extent of the decision-making process of the model. At a low level, we generate a candidate $g$ that encapsulates the approximations $f_{i,m_i}^* \tilde= \hat{f}_{i,m_i}$ of the behavior of certain functions learned by $M$ at some steps $s_i$. On an abstract level, interpretations can be seen as hypotheses about the role of evidence in the explanation-generation process. Like a good experimental hypothesis, a good interpretation satisfies two core properties: (i) is testable, and (ii) clearly defines dependent and independent variables. Interpretations can be formulated using different forms of reasoning (\textit{e.g.}, deductive, inductive, abductive, etc.). In particular, the survey on explanations and social sciences by Miller reports that people usually make assumptions (\textit{i.e.}, in our context, choose an interpretation) via social attribution of intent (to the evidence)~\cite{Miller2017-wj}. Social attribution is concerned with how people attribute or explain the behavior of others, and not with the real causes of the behavior. Social attribution is generally expressed through folk psychology, which is the attribution of intentional behavior using everyday terms such as beliefs, desires, intentions, emotions, and personality traits. Such concepts may not truly be the cause of the described behavior but are indeed those humans leverage to model and predict each others’ behaviors. This may generate misalignment between a hypothesized interpretation of some evidence and its actual role within the inference process of the model. In other words, reasoning on evidence through folk psychology might generate interpretations that are \emph{plausible} but not necessarily \emph{faithful} to the inference process of the model (such terms will be further explored in \autoref{sec:faithfulness-vs-plausibility}).

\subsection{Explanation Interface}
\label{sec:xui}

Explanations are meant to be delivered to some target users. We define \ac{XUI} as the format in which some explanation is presented to the end user. This could be, for example, in the form of text, plots, infographics, etc. We argue that an \ac{XUI} is characterized by three main properties: (i) human understandability, (ii) informativeness, and (iii) completeness. The \emph{human-understandability} is the degree to which users can understand the answer to their ``why'' question via the \ac{XUI}. This property depends on user cognition, bias, expertise, goals, etc., and is influenced by the complexity of the selected interpretation function. The \emph{informativeness} (i.e., depth) of an explanation is a measure of the effectiveness of an \ac{XUI} in answering the why question posed by the user. That is the depth of information for some $s_i$ of great interest in the \ac{XUI}. The \emph{completeness} (i.e., width) of an explanation is the accuracy of an \ac{XUI} in describing the overall model’s workings, and the degree to which it allows for anticipating predictions. That is the width in terms of the number of $s_i$ the \ac{XAI} spans. Note that both informativeness and completeness are bound by the explanatory potential of the evidence (\textit{e.g.}, attention weights do not explain the full model, just some transformation steps, while the full set of model parameters does). 

\section{Concerning Faithfulness and Plausibility}
\label{sec:faithfulness-vs-plausibility}

In \autoref{sec:generating-interpretations} we observed that social attribution is a double-edged sword for the interpretation generation process, as it may incur in propelling plausibility without accounting for faithfulness. This issue was highlighted by Jacovi \& Goldberg, who introduced a property of explanations called \emph{aligned faithfulness} \cite{Jacovi2021-pi}. In the words of the authors, an explanation satisfies this property if ``it is faithful and aligned to the social attribution of the intent behind the causal chain of decision-making processes". Our proposed framework allows us to go a step forward in the characterization of this property. We note that the property of aligned faithfulness pertains only to interpretations, not evidence. The latter by itself has no inherent meaning, its semantics is defined by some interpretation that may or may not involve social attribution of intent to the causal chain of inference processes.

\begin{figure}[t]
    \centering
    \includegraphics[width=0.45\textwidth]{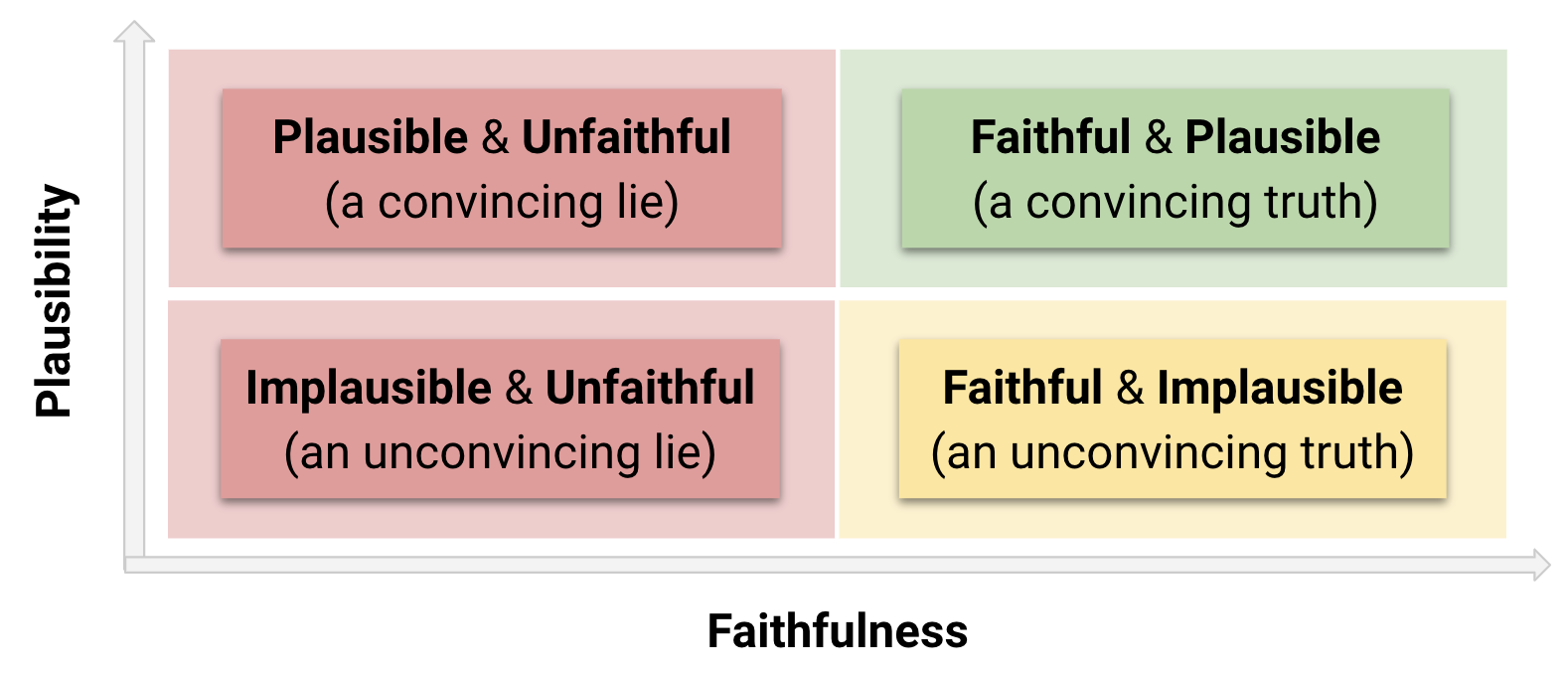}
    \caption{Overview of the outcome on the user of the interaction between faithfulness and plausibility.}
    \label{fig:plausibility-vs-faithfulness}
\end{figure}

\subsection{Faithfulness}
\label{sec:faithfulness}

Given an interpretation function $g$ describing some transformation steps $s_i$ within a model $M$’s inference process, we want to be able to prove that $g$ is faithful (at least to some extent) to the actual transformations made by $M$ to an input $x$ to get a prediction $\hat{y}$. Namely, we define the property of \emph{faithfulness of an interpretation} $\phi_i(g, e)$, as the extent to which an interpretation $g$ accurately describes the behavior of some transformation functions $f_{i,m_i}$ that some model learned to map an output $x_{i-1, j}$ at $s_{i-1}$ into $x_{i,z}$ at $s_i$ making use of some instance evidence $e$. Given some evidence $e$ and its interpretation function $g$, we say that a related explanation is faithful to some transformation steps if the following conditions hold: (i) the evidence $e$ has explanatory potential $\epsilon_i>0$, and (ii) the interpretation $g$ has faithfulness $\phi_i>0$. Then we can define the \emph{faithfulness of an explanation} ($\Phi$) as a function of the faithfulness of the interpretation of each step involved and the related explanatory potential. For example, we could define $\Phi = \sum_i \epsilon_i \phi_i ~\forall i \in I \subseteq [1, N]$ where $I$ is the set of indices of transformation steps $s_i$ that involved the evidence $e$. Thus the faithfulness of an explanation is the sum of the faithfulness scores of its components, \textit{i.e.}, the faithfulness of the interpretations of the evidence involved in the generation of the explanation. Besides, the related explanatory power weights the faithfulness of each interpretation, following the intuition that evidence with higher $\epsilon$ should have a larger impact on the overall faithfulness score of the interpretation.

We can have various measures of faithfulness that are associated with different explanation types, in the same way as we have different metrics to evaluate the ability of a \ac{ML} model to complete a task. Thus $\phi_i$ is implicitly bounded.

When designing a faithful explanatory method, we can opt for two approaches. We can achieve faithfulness ``structurally" by enforcing this property on pre-selected interpretations in model design (\textit{e.g.}, imposing constraints on transformation steps limiting the range of learnable functions). This direction has been recently explored by Jain et al.~\cite{Jain2020-rj} and Jacovi \& Goldberg~\cite{Jacovi2021-pi}. An alternative, naive, strategy is trial-and-error: formulating interpretations and assessing their faithfulness via formal proofs or requirements-based testing using proxy tasks. While formal proofs are still missing in current literature, a number of tests for faithfulness have been recently proposed~\cite{Adebayo2018-yp,Jain2019-oe,Wiegreffe2019-ht,Serrano2019-tm}.

\subsection{Plausibility}
\label{sec:plausibility}

The combined value of the three above-mentioned properties of \acp{XUI} (\textit{i.e.}, human understandability, informativeness, and completeness) drives the plausibility of an explanation. More specifically, we define \emph{plausibility} as the degree to which an explanation is aligned with the user’s understanding of the model's partial or overall inner workings. Plausibility is a user-dependent property and as such, it is subject to the user’s knowledge, bias, etc. Unlike faithfulness, the plausibility of explanations can be assessed via user studies. Note that a plausible explanation is not necessarily faithful, just like a faithful explanation is not necessarily plausible. It is desirable for both properties to be satisfied in the design of some explanation. Interestingly, an unfaithful but plausible explanation may deceive a user into believing that a model behaves according to a rationale when it is actually not the case. This raises ethical concerns around the possibility that poorly designed explanations could spread inaccurate or false knowledge among the end-users. \autoref{fig:plausibility-vs-faithfulness} provides a simplified overview of the problem.

\section{Framing Common Explainability Strategies}
\label{sec:case-studies}

\subsection{Attention}
\label{sec:attention}

The introduction of attention mechanisms has been one of the most notable breakthroughs in \ac{DL} research in recent years. Originally proposed for empowering neural machine translation tasks~\cite{Bahdanau2014-fv}, it is currently employed in many state-of-the-art approaches for numerous cognitive tasks. The chain of transformations in the simplest neural model making use of self-attention is a three-step causal process: (i) encoding, (ii) weight encodings by attention scores, and (iii) decoding into model output. Then we can define the function learned by the model as the composition $\hat{f} = f_{3, m_3} \circ f_{2, m_2} \circ f_{1, m_1}$, where each $f_i$ for $i\in [1,3]$ corresponds to the respective transformation function in the causal chain.

\noindent \textbf{Evidence.} For an input $x$ split into $t$ sequentially related tokens, let $f_{1, m_1}$ be an encoder function such that $f_{1, m_1}(x)=\bar{X}$ is the vector of the encoded model input tokens. Then, $f_{2, m_2}(\bar{X})=\sum_{j=1}^t \alpha_j \bar{x}_j$, for all model input tokens $\bar{x}_j \in \bar{X}$, is the linear combination of the encodings weighted by their corresponding attention scores. 
Then $e_{att} = \{\alpha_j\}_1^t$.

\begin{equation}
    e_{att} = \{\alpha_j ~| ~f_{2, m_2}(\bar{X}) = \sum_{j=1}^t \alpha_j \bar{x}_j\}
\end{equation}

That is, the evidence $e_{att}$ related to a model input is the set of weights $\alpha_j$ produced by the attention layer. The explanatory potential $\epsilon(e_{att})$ is the ratio between the number of parameters involved in the analyzed attention layer with respect to the total number of parameters of the model.

\noindent \textbf{Interpretation.} The interpretation of the evidence is a function $g_{att}(e(x,\hat{y}))$ that describes function $f_{3, m_3}$, \textit{i.e.}, how the weighted encodings are decoded into the model conclusion.

\noindent \textbf{Faithfulness.} Note that we do not know the faithful interpretation function, so we hypothesize its behavior by formulating a candidate interpretation, a process that is usually guided by the researcher's intuition. In the case of attention, an interpretation generally shared among researchers is that ``the value of each attention weight describes the importance of the corresponding token in the original input to the model output". Unfortunately, albeit plausible, research in this field disproved such an interpretation of attention weights \cite{Jain2019-oe,Wiegreffe2019-ht,Serrano2019-tm}, leaving the role of attention for explainability (if any) still unclear.

\subsection{Grad-CAM}
\label{sec:gradient}
A popular explanation called \ac{Grad-CAM}~\cite{Selvaraju2016-hw} presents a method to explain a prediction made by an image classifier using the information encompassed in the back-propagated gradient of a prediction.
In short, \ac{Grad-CAM} uses the information about the gradient computed at the last convolutional layer of a \ac{CNN} given a certain input $x$ to assign a feature importance score for each
input feature.

\noindent \textbf{Evidence.} The \ac{Grad-CAM} evidence-extraction $\xi_{grad}$ method consisted of using the feature activation map of a convolutional layer from a given input $x$ to compute the neurons' importance weights $\alpha_{i}$. The explanatory potential $\epsilon(e_{grad})$ is related, as for the attention mechanism, to the number of parameters analyzed w.r.t the total number of parameters of the method. 

\noindent \textbf{Interpretation.} \ac{Grad-CAM} claim that the computed neuron's weights $\alpha_i$ corresponds to the part of the input features that influence the final prediction the most.

\noindent \textbf{Faithfulness.} The authors measure the faithfulness of the model using image occlusion. That is, they patched some part of the input to the model, and they measured the correlation with the difference in the final output. With this faithfulness metric, a high correlation means a high faithfulness in the explanation. 

\subsection{SHAP}
\label{sec:shap}

Lundberg \& Lee in 2017 proposed \ac{SHAP}~\cite{Lundberg2017-ar}, a method to assign an importance value to each feature used by an opaque model $M$ to explain a single prediction $\hat{y}$.
\ac{SHAP} has been presented as a generalization of other well-known explanation methods, such as \ac{LIME}~\cite{Ribeiro2016-wr}, DeepLIFT~\cite{Shrikumar2017-si}, Layer-wise relevance propagation~\cite{Bach2015-np}, and classic Shapley value estimation~\cite{Lundberg2017-ar}.
The \ac{SHAP} values are defined as:

\begin{equation}
   \label{eq:shap_equation}
   h\left(z^{\prime}\right)=\beta_0+\sum_{i=1}^M \beta_i z_i^{\prime}
\end{equation}

\noindent where $z_i^{\prime} \in \{0,1\}^M$ is a simplified version of the input $x$, $M$ is the number of features used in the explanation, and $\beta_i \in \mathbb{R}$ is a coefficient that represents the effect that the $i-th$ feature has on the output.

\noindent \textbf{Evidence.} The only evidence $e_{shape} = \xi_{shap}(M, x)$ used by \ac{SHAP} is the set of predictions made by the classifier in a neighborhood of $x$.
To compute the explanatory potential $\epsilon\left({e_{shap}}\right)$, we can use the ratio of predictions employed to compute the \ac{SHAP} values w.r.t the total number of possible samples in the countable (and possibly infinite) neighborhood of $x$. Thus, the greater the number of predictions we have, the higher the exploratory potential of the method.

\noindent \textbf{Interpretation.} The interpretation $g_{shap}$ of the evidence proposed by \ac{SHAP} is that, given $e_{shap}$, we can locally reproduce the behavior of a complex unknown model with a simple additive model $h\left( \cdot \right)$, and analyzing $h\left( \cdot \right)$ we can get a local explanation $E_{shap}$ of the behavior of the initial model. 
That is, the proposed interpretation of the evidence results from the optimization problem in~\autoref{eq:shap_equation}.

\noindent \textbf{Faithfulness.}
Even though the authors do not present a measure of the faithfulness of the explanation directly, they provide three desirable  properties that are \emph{i)} local accuracy, \emph{ii)} missingness, \emph{iii)} consistency.
The authors showed that their method is the only one that satisfies all these properties, assessing a requirements-based form of faithfulness as described in §~\ref{sec:faithfulness-vs-plausibility}.

\subsection{Linear regression models}
\label{sec:linear_regression}

Linear regression models are not an explanation method but are normally considered \emph{intrinsically interpretable}.
Following our proposed framework, we claim that defining them, among other models, as \emph{intrinsically interpretable} is inaccurate and often misleading.
In fact, the definition of what is simple to be interpreted by humans is not well-defined, and we can enumerate various examples of models that are easy to be interpreted by a practitioner but are almost black-boxed for non-expert users.

A linear regressor $\hat{f}_{lin}(\cdot)$ is typically formulated as:
\begin{equation}
   \label{eq:linear_reg}
   \hat{f}_{lin}(x)=\beta_0+\sum_{i=1}^N \beta_i x_i^{\prime}
\end{equation}
where $\beta_i$ are the weights of the learned features, and $N$ is the feature space dimension.

\noindent \textbf{Evidence.} The implicit assumption, claiming that a linear model is intrinsically interpretable, is that the weights ${\beta_i, 1 \leq i \leq M}$ are a good explanation for the model.
Thus $e_{lin} = \{\beta_i\}_1^N$.
With a linear model, we have the maximum explanatory potential $\xi_{lin}$ because with $e_{lin}$ we can fully describe the model.

\noindent \textbf{Interpretation.} Assuming a normalization of the features, we can say that the higher the value of $\beta_i$, the higher the contribution of the feature $x_i$ to the model prediction.

\noindent \textbf{Faithfulness.} There are no doubts about the faithfulness of the interpretation of the predictions given the normalization assumption, and in fact, a linear model is normally considered an intrinsically interpretable method. However, in a real scenario, its plausibility to a non-expert user is not guaranteed. 

\subsection{Fuzzy models}
\label{sec:fis}

Fuzzy models, especially in the form of \ac{FRBS}, represent effective tools for the modeling of complex systems by using a human-comprehensible linguistic approach. 
Thanks to these characteristics, they are generally considered white or gray boxes and are often mentioned as good options for interpretable AI~\cite{fuchs2022impact}. 
\acp{FRBS} perform their inference (i.e., calculate a conclusion) by exploiting a knowledge base composed of linguistic terms and rules. 
\noindent
A fuzzy rule is usually expressed as a sentence in the form:

\begin{equation}
\texttt{IF <antecedent> THEN <consequent>}
\end{equation}

\noindent
where \texttt{antecedent} is a logic formula created by concatenating clauses like \texttt{`X IS a'} with some logical operators, where \texttt{T} is a linguistic variable (associated with one input feature) and $a$ is a linguistic term.
Thanks to this representation, the antecedent of each rule give an intuitive and human-understandable characterization of some class/group.

\noindent \textbf{Evidence.} The rules are good evidence for a large part of the model: they characterize the feature space by using a self-explanatory formalism that can be read and validated, by human operators.

\noindent \textbf{Interpretation.} The fuzzy sets, that are used to create the fuzzy terms and evaluate the satisfaction of the antecedents, have self-explanatory interpretations: they define how much a value belongs to a given set by means of membership functions. The fuzzy rules are also self-explanatory. The only part that requires a proper interpretation is the output calculation function. In the case of Sugeno reasoning, such functions can be seen as linear regression models, hence all considerations discussed in \autoref{sec:linear_regression} remain valid also in the case of fuzzy models. 

\noindent \textbf{Faithfulness.}
Similarly to the case of linear regression models, there are no doubts about the faithfulness of the interpretation of the predictions given a normalization step. However, in the case of special transformations (e.g., log-transformation), some of the intrinsic interpretability might be lost in favor of better fitting to training data \cite{fuchs2022impact}.  
Since it is often the case that features in biomedicine (see, e.g., clinical parameters) follow  a log-normal distribution, such transformations are very frequent and delicate.

\section{On the impact on biomedicine}

AI models have revolutionized the field of biomedicine, enabling advanced analyses, predictions, and decision-making processes. However, the increasing complexity of AI models, such as deep learning neural networks, has raised concerns regarding their lack of explainability. The present paper provides a common ground on theoretical notions of explainability, as a pre-requisite to a principled discussion and evaluation. The final aim is to catalyze a coherent examination of the impact of AI and explainability by highlighting its significance in facilitating trust, regulatory compliance, and accelerating research and development. Such a topic is particularly well-suited for high-stake environments, such as biomedicine.

The use of black box machine learning models in the biomedical field has been steadily increasing, with many researchers relying on them to make predictions and gain insights from complex datasets. However, the opacity of these models poses a challenge to their explainability, leading to the question of what criteria should be used to evaluate their explanations \cite{Combi2022-oo}. Two key criteria that have been proposed are \textit{faithfulness} and \textit{plausibility}. Faithfulness is important because it allows researchers to understand how the model arrived at its conclusions, identify potential sources of error or bias, and ultimately increase the trustworthiness of the model. In the biomedical field, this is particularly important as the stakes are high when it comes to making accurate predictions about patient outcomes and the decision maker wants to have the most accurate motivation behind the model suggestions. On the other hand, while a faithful explanation may accurately reflect the model's inner workings, it may not be understandable or useful to stakeholders who lack expertise in the technical details of the model. Plausibility is therefore important because all the people involved in the decision-making process gain insights from the model that they can act upon and can communicate these insights to other stakeholders in a way that is meaningful and actionable. Both faithfulness and plausibility are important criteria for evaluating explanations of black-box machine-learning models in the biomedical field. Researchers should strive to balance these criteria to provide the best explanations to the stakeholders involved; this can be achieved by involving them during the entire development of the AI machine models and tools.

Beyond the need for faithfulness and plausibility, the concrete impact of explainability on AI for biomedicine is broad. Explainability plays a crucial role in establishing trust between healthcare professionals and AI systems. In critical biomedical applications, such as disease diagnosis, treatment recommendation, and patient monitoring, transparency in decision-making is essential for clinicians to make informed decisions. By providing interpretable explanations, AI models can help healthcare professionals understand the reasoning behind the model's predictions, leading to increased trust and acceptance \cite{10.1145/2783258.2788613}. Moreover, regulatory bodies of biomedicine, such as the Food and Drug Administration (FDA) in the United States, require transparency and accountability for AI models used in clinical decision-making. XAI techniques provide an opportunity to meet regulatory standards by enabling model auditing and validation. Through explainability, clinicians and regulatory bodies can assess the risk associated with the deployment of AI models, ensuring patient safety and compliance with ethical guidelines \cite{10.1145/3465416.3483305}. Finally, explainability can significantly enhance the research and development process in biomedicine. By uncovering the underlying factors and features that contribute to an AI model's decision, researchers can gain valuable insights into disease mechanisms, biomarkers, and potential therapeutic targets.

\section{Conclusions and Future Work}
\label{sec:conclusions}

In this work, we propose a novel theoretical framework that brings order and opportunities for a better design of explanations to the \ac{XAI} community by introducing formal terminology. The framework allows dissecting explanations into evidence (factual data coming from the model) and interpretation (a hypothesized function that describes how the model uses the evidence). The explanation is the product of the application of the interpretation to the evidence and is presented to the target user via some form of explanation interface. 
These components allow for designing more principled explanations by defining the atomic components and the properties that enable them. There are three core properties: \emph{(i)} the explanatory potential for the evidence (i.e., how much of the model the evidence can tell about); \emph{(ii)} the faithfulness of the interpretation (i.e., whether the interpretation is true to the decision-making of the model); \emph{(iii)} the plausibility of the explanation interface (i.e., how much the explanation makes sense to the user and is intelligible). 
We show that the theoretical framework can be applied to explanations coming from a variety of methods, which fit the atomic components we propose. 
The lesson learned from analyzing explanations over the lent of our proposed framework is that humans (both stakeholder and researcher) should be involved in the design of explainability as soon as possible in the \ac{AI}-powered software design process, especially in sensitive application domains like biomedicine, where a blind application of black-box approaches hampers the right to an explanation. 
Involving stakeholders allows for a proper filling of each component in the theoretical framework of explainability, and informs model design. 
The top-down approach that is established this way propels the human understanding of how \ac{AI} (and \ac{ML} in particular) works, possibly fostering user trust in the system. We believe that high stake decision-making domains such as biomedicine would be those which will benefit the most from a more rigorous definition of core concepts of explainability, with opportunities to cement a conscious aid of AI-assisted decisions. Therefore, in future work, we want to apply our theoretical study to a real-case scenario in the biomedical sector and analyze its implementation with the help of human feedback, to better focus on the plausibility analysis of our theoretical framework.

\balance

\bibliographystyle{IEEEtran}
\bibliography{paper}

\begin{thebibliography}{10}
\providecommand{\url}[1]{#1}
\csname url@samestyle\endcsname
\providecommand{\newblock}{\relax}
\providecommand{\bibinfo}[2]{#2}
\providecommand{\BIBentrySTDinterwordspacing}{\spaceskip=0pt\relax}
\providecommand{\BIBentryALTinterwordstretchfactor}{4}
\providecommand{\BIBentryALTinterwordspacing}{\spaceskip=\fontdimen2\font plus
\BIBentryALTinterwordstretchfactor\fontdimen3\font minus
  \fontdimen4\font\relax}
\providecommand{\BIBforeignlanguage}[2]{{%
\expandafter\ifx\csname l@#1\endcsname\relax
\typeout{** WARNING: IEEEtran.bst: No hyphenation pattern has been}%
\typeout{** loaded for the language `#1'. Using the pattern for}%
\typeout{** the default language instead.}%
\else
\language=\csname l@#1\endcsname
\fi
#2}}
\providecommand{\BIBdecl}{\relax}
\BIBdecl

\bibitem{Rudin2019-bj}
C.~Rudin, ``\BIBforeignlanguage{en}{Stop explaining black box machine learning
  models for high stakes decisions and use interpretable models instead},''
  \emph{\BIBforeignlanguage{en}{Nat Mach Intell}}, vol.~1, no.~5, pp. 206--215,
  May 2019.

\bibitem{Kundu2021}
S.~Kundu, ``Ai in medicine must be explainable,'' \emph{Nature Medicine},
  vol.~27, no.~8, pp. 1328--1328, Aug 2021.

\bibitem{Goodman2017-iw}
B.~Goodman and S.~Flaxman, ``\BIBforeignlanguage{en}{European union regulations
  on algorithmic {Decision-Making} and a ``right to explanation''},''
  \emph{\BIBforeignlanguage{en}{AIMag}}, vol.~38, no.~3, pp. 50--57, Oct. 2017.

\bibitem{Chen2018-eq}
C.~Chen, O.~Li, C.~Tao, A.~J. Barnett, J.~Su, and C.~Rudin, ``This looks like
  that: Deep learning for interpretable image recognition,'' in
  \emph{Proceedings of the 33rd International Conference on Neural Information
  Processing Systems}.\hskip 1em plus 0.5em minus 0.4em\relax Red Hook, NY,
  USA: Curran Associates Inc., 2019.

\bibitem{Zhang2017-by}
Q.~Zhang, Y.~N. Wu, and S.-C. Zhu, ``Interpretable convolutional neural
  networks,'' in \emph{2018 IEEE/CVF Conference on Computer Vision and Pattern
  Recognition}, 2018, pp. 8827--8836.

\bibitem{Hou2020-zf}
B.-J. Hou and Z.-H. Zhou, ``\BIBforeignlanguage{en}{Learning with interpretable
  structure from gated {RNN}},'' \emph{\BIBforeignlanguage{en}{IEEE Trans
  Neural Netw Learn Syst}}, vol.~31, no.~7, pp. 2267--2279, Jul. 2020.

\bibitem{Ribeiro2016-uy}
M.~T. Ribeiro, S.~Singh, and C.~Guestrin, ````why should {I} trust you?'':
  Explaining the predictions of any classifier,'' in \emph{Proceedings of the
  22nd {ACM} {SIGKDD} International Conference on Knowledge Discovery and Data
  Mining}, ser. KDD '16.\hskip 1em plus 0.5em minus 0.4em\relax New York, NY,
  USA: Association for Computing Machinery, Aug. 2016, pp. 1135--1144.

\bibitem{Lundberg2017-ar}
S.~M. Lundberg and S.-I. Lee, ``A unified approach to interpreting model
  predictions,'' in \emph{Proceedings of the 31st International Conference on
  Neural Information Processing Systems}, ser. NIPS'17.\hskip 1em plus 0.5em
  minus 0.4em\relax Red Hook, NY, USA: Curran Associates Inc., 2017, p.
  4768–4777.

\bibitem{Bahdanau2014-fv}
\BIBentryALTinterwordspacing
D.~Bahdanau, K.~Cho, and Y.~Bengio, ``Neural machine translation by jointly
  learning to align and translate,'' in \emph{3rd International Conference on
  Learning Representations, {ICLR} 2015, San Diego, CA, USA, May 7-9, 2015,
  Conference Track Proceedings}, Y.~Bengio and Y.~LeCun, Eds., 2015. [Online].
  Available: \url{http://arxiv.org/abs/1409.0473}
\BIBentrySTDinterwordspacing

\bibitem{Jain2019-oe}
S.~Jain and B.~C. Wallace, ``\BIBforeignlanguage{en}{Attention is not
  explanation},'' in \emph{\BIBforeignlanguage{en}{Proceedings of the 2019
  Conference of the North}}.\hskip 1em plus 0.5em minus 0.4em\relax
  Stroudsburg, PA, USA: Association for Computational Linguistics, 2019.

\bibitem{Wiegreffe2019-ht}
S.~Wiegreffe and Y.~Pinter, ``Attention is not not explanation,'' in
  \emph{Proceedings of the 2019 Conference on Empirical Methods in Natural
  Language Processing and the 9th International Joint Conference on Natural
  Language Processing ({{EMNLP-IJCNLP}})}.\hskip 1em plus 0.5em minus
  0.4em\relax Stroudsburg, PA, USA: Association for Computational Linguistics,
  2019, pp. 11--20.

\bibitem{Serrano2019-tm}
S.~Serrano and N.~A. Smith, ``Is attention interpretable?'' in
  \emph{Proceedings of the 57th Annual Meeting of the Association for
  Computational Linguistics}.\hskip 1em plus 0.5em minus 0.4em\relax Florence,
  Italy: Association for Computational Linguistics, Jul. 2019, pp. 2931--2951.

\bibitem{Graziani}
M.~Graziani, L.~Dutkiewicz, D.~Calvaresi, J.~Amorim, K.~Yordanova, M.~Vered,
  R.~Nair, P.~Henriques~Abreu, T.~Blanke, V.~Pulignano, J.~Prior, L.~Lauwaert,
  W.~Reijers, A.~Depeursinge, V.~Andrearczyk, and H.~Müller, ``A global
  taxonomy of interpretable ai: unifying the terminology for the technical and
  social sciences,'' \emph{Artificial Intelligence Review}, 09 2022.

\bibitem{clinciu-hastie-2019-survey}
\BIBentryALTinterwordspacing
M.-A. Clinciu and H.~Hastie, ``A survey of explainable {AI} terminology,'' in
  \emph{Proceedings of the 1st Workshop on Interactive Natural Language
  Technology for Explainable Artificial Intelligence (NL4XAI 2019)}.\hskip 1em
  plus 0.5em minus 0.4em\relax Association for Computational Linguistics, 2019,
  pp. 8--13. [Online]. Available: \url{https://aclanthology.org/W19-8403}
\BIBentrySTDinterwordspacing

\bibitem{murdoch_definitions_2019}
\BIBentryALTinterwordspacing
W.~J. Murdoch, C.~Singh, K.~Kumbier, R.~Abbasi-Asl, and B.~Yu,
  ``\BIBforeignlanguage{EN}{Definitions, methods, and applications in
  interpretable machine learning},'' \emph{\BIBforeignlanguage{EN}{Proceedings
  of the National Academy of Sciences}}, vol. 116, no.~44, pp.
  22\,071--22\,080, Oct. 2019, company: National Academy of Sciences
  Distributor: National Academy of Sciences Institution: National Academy of
  Sciences Label: National Academy of Sciences Publisher: Proceedings of the
  National Academy of Sciences. [Online]. Available:
  \url{https://www.pnas.org/doi/abs/10.1073/pnas.1900654116}
\BIBentrySTDinterwordspacing

\bibitem{Arbelaez_Ossa2022-ve}
L.~Arbelaez~Ossa, G.~Starke, G.~Lorenzini, J.~E. Vogt, D.~M. Shaw, and B.~S.
  Elger, ``\BIBforeignlanguage{en}{Re-focusing explainability in medicine},''
  \emph{\BIBforeignlanguage{en}{Digit Health}}, vol.~8, p. 20552076221074488,
  Feb. 2022.

\bibitem{Jacovi2020-ec}
\BIBentryALTinterwordspacing
A.~Jacovi and Y.~Goldberg, ``Towards faithfully interpretable {NLP} systems:
  How should we define and evaluate faithfulness?'' in \emph{Proceedings of the
  58th Annual Meeting of the Association for Computational Linguistics}.\hskip
  1em plus 0.5em minus 0.4em\relax Online: Association for Computational
  Linguistics, Jul. 2020, pp. 4198--4205. [Online]. Available:
  \url{https://aclanthology.org/2020.acl-main.386}
\BIBentrySTDinterwordspacing

\bibitem{Jacovi2021-pi}
------, ``\BIBforeignlanguage{en}{Aligning faithful interpretations with their
  social attribution},'' \emph{\BIBforeignlanguage{en}{Trans. Assoc. Comput.
  Linguist.}}, vol.~9, pp. 294--310, Mar. 2021.

\bibitem{Vaswani2017-kq}
\BIBentryALTinterwordspacing
A.~Vaswani, N.~Shazeer, N.~Parmar, J.~Uszkoreit, L.~Jones, A.~N. Gomez, L.~u.
  Kaiser, and I.~Polosukhin, ``Attention is all you need,'' in \emph{Advances
  in Neural Information Processing Systems}, I.~Guyon, U.~V. Luxburg,
  S.~Bengio, H.~Wallach, R.~Fergus, S.~Vishwanathan, and R.~Garnett, Eds.,
  vol.~30.\hskip 1em plus 0.5em minus 0.4em\relax Curran Associates, Inc.,
  2017. [Online]. Available:
  \url{https://proceedings.neurips.cc/paper/2017/file/3f5ee243547dee91fbd053c1c4a845aa-Paper.pdf}
\BIBentrySTDinterwordspacing

\bibitem{Selvaraju2016-hw}
R.~R. Selvaraju, M.~Cogswell, A.~Das, R.~Vedantam, D.~Parikh, and D.~Batra,
  ``Grad-cam: Visual explanations from deep networks via gradient-based
  localization,'' in \emph{2017 IEEE International Conference on Computer
  Vision (ICCV)}, 2017, pp. 618--626.

\bibitem{Mittelstadt2019-jk}
B.~Mittelstadt, C.~Russell, and S.~Wachter, ``Explaining explanations in
  {AI},'' in \emph{Proceedings of the Conference on Fairness, Accountability,
  and Transparency}, ser. FAT* '19.\hskip 1em plus 0.5em minus 0.4em\relax New
  York, NY, USA: Association for Computing Machinery, Jan. 2019, pp. 279--288.

\bibitem{Molnar2022-kh}
\BIBentryALTinterwordspacing
C.~Molnar, \emph{Interpretable Machine Learning}, 2nd~ed.\hskip 1em plus 0.5em
  minus 0.4em\relax Independently published (February 28, 2022), 2022.
  [Online]. Available: \url{https://christophm.github.io/interpretable-ml-book}
\BIBentrySTDinterwordspacing

\bibitem{Chazette2020-zz}
L.~Chazette and K.~Schneider, ``Explainability as a non-functional requirement:
  challenges and recommendations,'' \emph{Requirements Engineering}, vol.~25,
  no.~4, pp. 493--514, Dec. 2020.

\bibitem{Li2022-xt}
B.~Li, P.~Qi, B.~Liu, S.~Di, J.~Liu, J.~Pei, J.~Yi, and B.~Zhou, ``Trustworthy
  {AI}: From principles to practices,'' \emph{ACM Comput. Surv.}, Aug. 2022.

\bibitem{Lipton2016-ba}
Z.~C. Lipton, ``The mythos of model interpretability: In machine learning, the
  concept of interpretability is both important and slippery.'' \emph{Queue},
  vol.~16, no.~3, p. 31–57, jun 2018.

\bibitem{Guidotti2018-ti}
R.~Guidotti, A.~Monreale, S.~Ruggieri, F.~Turini, F.~Giannotti, and
  D.~Pedreschi, ``A survey of methods for explaining black box models,''
  \emph{ACM Comput. Surv.}, vol.~51, no.~5, aug 2018.

\bibitem{Murdoch2019-wk}
W.~J. Murdoch, C.~Singh, K.~Kumbier, R.~Abbasi-Asl, and B.~Yu,
  ``\BIBforeignlanguage{en}{Definitions, methods, and applications in
  interpretable machine learning},'' \emph{\BIBforeignlanguage{en}{Proc. Natl.
  Acad. Sci. U. S. A.}}, vol. 116, no.~44, pp. 22\,071--22\,080, Oct. 2019.

\bibitem{Mueller2019-jr}
\BIBentryALTinterwordspacing
S.~T. Mueller, R.~R. Hoffman, W.~J. Clancey, A.~Emrey, and G.~Klein,
  ``Explanation in human-ai systems: {A} literature meta-review, synopsis of
  key ideas and publications, and bibliography for explainable {AI},''
  \emph{CoRR}, vol. abs/1902.01876, 2019. [Online]. Available:
  \url{http://arxiv.org/abs/1902.01876}
\BIBentrySTDinterwordspacing

\bibitem{Miller2017-wj}
\BIBentryALTinterwordspacing
T.~Miller, ``Explanation in artificial intelligence: Insights from the social
  sciences,'' \emph{Artif. Intell.}, vol. 267, pp. 1--38, 2019. [Online].
  Available: \url{https://doi.org/10.1016/j.artint.2018.07.007}
\BIBentrySTDinterwordspacing

\bibitem{Jain2020-rj}
\BIBentryALTinterwordspacing
S.~Jain, S.~Wiegreffe, Y.~Pinter, and B.~C. Wallace, ``{L}earning to faithfully
  rationalize by construction,'' in \emph{Proceedings of the 58th Annual
  Meeting of the Association for Computational Linguistics}.\hskip 1em plus
  0.5em minus 0.4em\relax Online: Association for Computational Linguistics,
  Jul. 2020, pp. 4459--4473. [Online]. Available:
  \url{https://aclanthology.org/2020.acl-main.409}
\BIBentrySTDinterwordspacing

\bibitem{Adebayo2018-yp}
J.~Adebayo, J.~Gilmer, M.~Muelly, I.~Goodfellow, M.~Hardt, and B.~Kim, ``Sanity
  checks for saliency maps,'' in \emph{Proceedings of the 32nd International
  Conference on Neural Information Processing Systems}, ser. NIPS'18.\hskip 1em
  plus 0.5em minus 0.4em\relax Red Hook, NY, USA: Curran Associates Inc., Dec.
  2018, pp. 9525--9536.

\bibitem{Ribeiro2016-wr}
\BIBentryALTinterwordspacing
M.~Ribeiro, S.~Singh, and C.~Guestrin, ``{``}why should {I} trust you?{''}:
  Explaining the predictions of any classifier,'' in \emph{Proceedings of the
  2016 Conference of the North {A}merican Chapter of the Association for
  Computational Linguistics: Demonstrations}.\hskip 1em plus 0.5em minus
  0.4em\relax San Diego, California: Association for Computational Linguistics,
  Jun. 2016, pp. 97--101. [Online]. Available:
  \url{https://aclanthology.org/N16-3020}
\BIBentrySTDinterwordspacing

\bibitem{Shrikumar2017-si}
A.~Shrikumar, P.~Greenside, and A.~Kundaje, ``Learning important features
  through propagating activation differences,'' in \emph{Proceedings of the
  34th International Conference on Machine Learning}, ser. Proceedings of
  Machine Learning Research, D.~Precup and Y.~W. Teh, Eds., vol.~70.\hskip 1em
  plus 0.5em minus 0.4em\relax Sydney, NSW, Australia: PMLR, 2017, pp.
  3145--3153.

\bibitem{Bach2015-np}
S.~Bach, A.~Binder, G.~Montavon, F.~Klauschen, K.-R. M{\"u}ller, and W.~Samek,
  ``\BIBforeignlanguage{en}{On {Pixel-Wise} explanations for {Non-Linear}
  classifier decisions by {Layer-Wise} relevance propagation},''
  \emph{\BIBforeignlanguage{en}{PLoS One}}, vol.~10, no.~7, p. e0130140, Jul.
  2015.

\bibitem{fuchs2022impact}
C.~Fuchs, S.~Spolaor, U.~Kaymak, and M.~S. Nobile, ``The impact of variable
  selection and transformation on the interpretability and accuracy of fuzzy
  models,'' in \emph{2022 IEEE Conference on Computational Intelligence in
  Bioinformatics and Computational Biology (CIBCB)}.\hskip 1em plus 0.5em minus
  0.4em\relax IEEE, 2022, pp. 1--8.

\bibitem{Combi2022-oo}
C.~Combi, B.~Amico, R.~Bellazzi, A.~Holzinger, J.~H. Moore, M.~Zitnik, and
  J.~H. Holmes, ``\BIBforeignlanguage{en}{A manifesto on explainability for
  artificial intelligence in medicine},'' \emph{\BIBforeignlanguage{en}{Artif
  Intell Med}}, vol. 133, p. 102423, Oct. 2022.

\bibitem{10.1145/2783258.2788613}
\BIBentryALTinterwordspacing
R.~Caruana, Y.~Lou, J.~Gehrke, P.~Koch, M.~Sturm, and N.~Elhadad,
  ``Intelligible models for healthcare: Predicting pneumonia risk and hospital
  30-day readmission,'' in \emph{Proceedings of the 21th ACM SIGKDD
  International Conference on Knowledge Discovery and Data Mining}, ser. KDD
  '15.\hskip 1em plus 0.5em minus 0.4em\relax New York, NY, USA: Association
  for Computing Machinery, 2015, p. 1721–1730. [Online]. Available:
  \url{https://doi.org/10.1145/2783258.2788613}
\BIBentrySTDinterwordspacing

\bibitem{10.1145/3465416.3483305}
\BIBentryALTinterwordspacing
H.~Suresh and J.~Guttag, \emph{A Framework for Understanding Sources of Harm
  throughout the Machine Learning Life Cycle}.\hskip 1em plus 0.5em minus
  0.4em\relax New York, NY, USA: Association for Computing Machinery, 2021.
  [Online]. Available: \url{https://doi.org/10.1145/3465416.3483305}
\BIBentrySTDinterwordspacing

\end{thebibliography}

\end{document}